\documentclass[conference]{IEEEtran}
\IEEEoverridecommandlockouts
\usepackage{cite}
\usepackage{amsmath,amssymb,amsfonts}
\usepackage{algorithmic}
\usepackage{graphicx}
\usepackage{textcomp}
\usepackage{xcolor}
\def\BibTeX{{\rm B\kern-.05em{\sc i\kern-.025em b}\kern-.08em
    T\kern-.1667em\lower.7ex\hbox{E}\kern-.125emX}}
\begin{document}

\title{Combating Uncertainty and Class Imbalance in Facial Expression Recognition\\
}

\author{
\IEEEauthorblockN{
Jiaxiang Fan\IEEEauthorrefmark{1},
Jian Zhou\IEEEauthorrefmark{1},
Xiaoyu Deng\IEEEauthorrefmark{1},
Huabin Wang\IEEEauthorrefmark{1},
Liang Tao\IEEEauthorrefmark{1} and
Hon Keung Kwan\IEEEauthorrefmark{2}}
\IEEEauthorblockA{\IEEEauthorrefmark{1}Anhui Provincial Key Laboratary of Multimodal Cognitive Computation, Anhui University, Hefei, Anhui, China 230601}
\IEEEauthorblockA{\IEEEauthorrefmark{2}Department of Electrical and Computer Engineering, University of Windsor, Windsor, Ontario, Canada N9B 3P4}
\IEEEauthorblockA{ \{e20201071@stu.ahu.edu.cn, jzhou@ahu.edu.cn, xy.ahu@qq.com, wanghuabin@ahu.edu.cn, taoliang@ahu.edu.cn,}
\IEEEauthorblockA{kwan1@uwindsor.ca\}}}

\maketitle

\begin{abstract}
Recognition of facial expression is a challenge when it comes to computer vision. The primary reasons are class imbalance due to data collection and uncertainty due to inherent noise such as fuzzy facial expressions and inconsistent labels. However, current research has focused either on the problem of class imbalance or on the problem of uncertainty, ignoring the intersection of how to address these two problems. Therefore, in this paper, we propose a framework based on Resnet and Attention to solve the above problems. We design weight for each class. Through the penalty mechanism, our model will pay more attention to the learning of small samples during training, and the resulting decrease in model accuracy can be improved by a Convolutional Block Attention Module (CBAM). Meanwhile, our backbone network will also learn an uncertain feature for each sample. By mixing uncertain features between samples, the model can better learn those features that can be used for classification, thus suppressing uncertainty. Experiments show that our method surpasses most basic methods in terms of accuracy on facial expression data sets (e.g., AffectNet, RAF-DB), and it also solves the problem of class imbalance well.
\end{abstract}


\section{Introduction}

\begin{figure}[htbp]
\centerline{\includegraphics[scale=0.54]{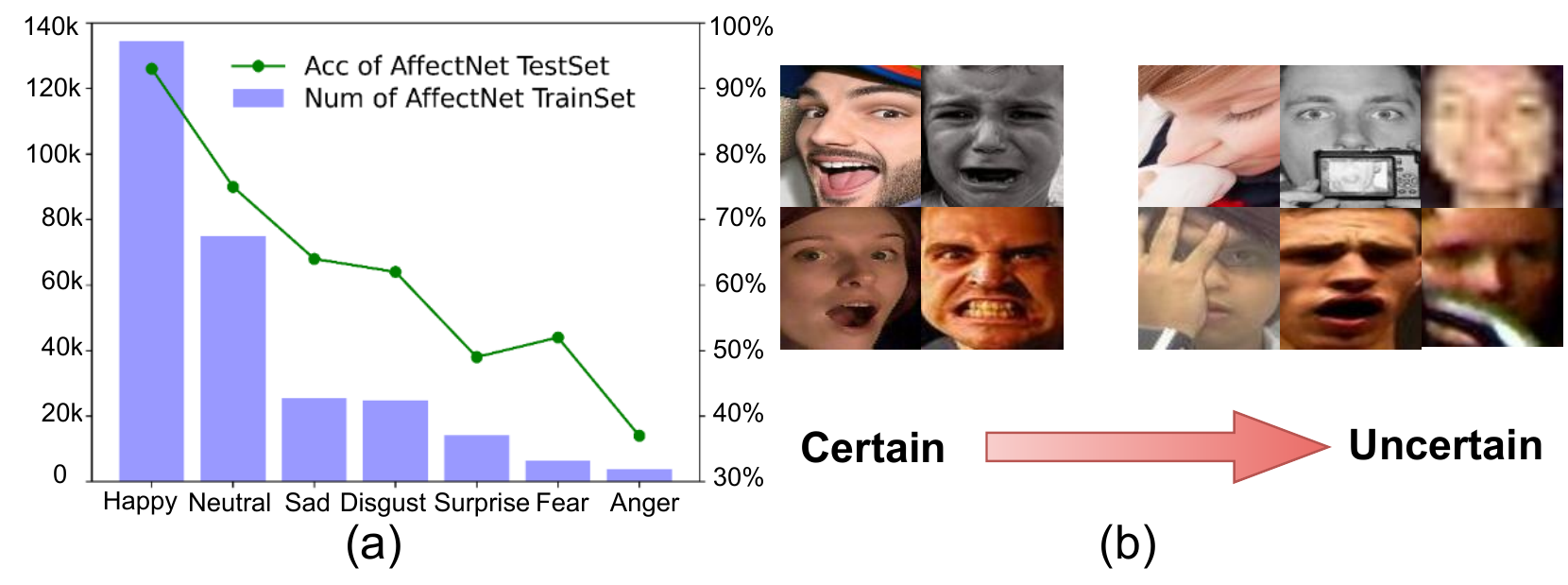}}
\caption{(a) The class distribution of FER training data is biased, resulting in different test accuracy among the classes. (b) Illustration of uncertainties on real-world facial images from RAF-DB.}
\label{fig1}
\end{figure}

Facial expressions are one of the most commonly used signals to express emotions and intentions\cite{tian2001recognizing}. However, as shown in Fig.~\ref{fig1}(a), the existing in-the-wild FER training datasets are almost biased towards some majority classes, which leads to poor test accuracy for classes with smaller samples. Deep neural networks trained with biased data tend to support large sample classes but perform poorly in small sample classes. Also, using deep learning methods to train networks requires large amounts of accurately labeled data. However, for large-scale FER datasets collected from the Internet, the uncertainty caused by the subjectivity of the annotators and the ambiguity of wild-face images pose a great difficulty for high-quality annotation. As shown in Fig.~\ref{fig1}(b), the uncertainties gradually change from high-quality and clear facial expressions to low-quality and blurred expressions. These uncertainties usually lead to inconsistencies between the real expressions and the labels of the samples, which seriously hinders the progress of facial expression recognition.

\begin{figure*}[htbp]
\centerline{\includegraphics[scale=0.7]{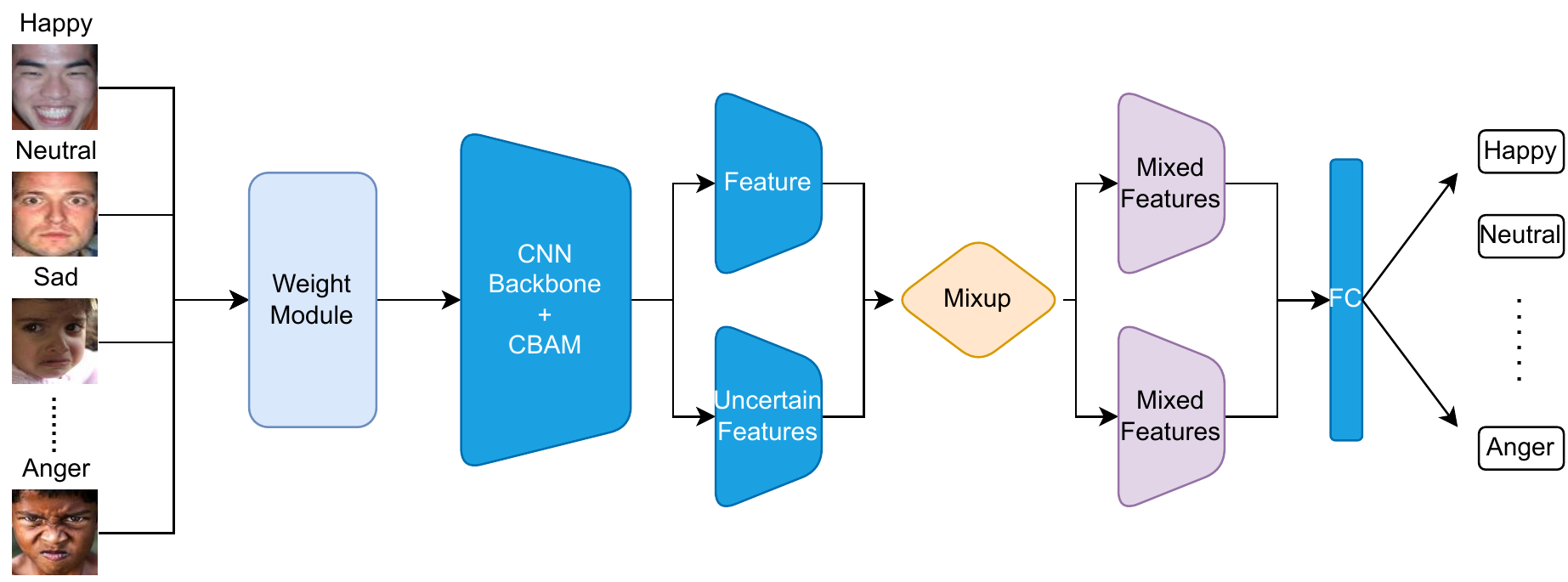}}
\caption{An overview of our proposed CUCN, before the pictures are input into the backbone network, we assign weights to them, and then embed CBAM in the backbone network.}
\label{fig2}
\end{figure*}

Recent studies have proposed various approaches to address the class imbalance and uncertainty in the field of facial expression recognition. Rifai proposed to enhance FER using small-scale controlled unlabeled data to improve performance and reduce uncertainty\cite{rifai2012disentangling}. In addition, some methods  remove class imbalance by transfer learning or meta-learning and learn domain invariant features of multiple datasets to recognize expressions. Learning the uncertainty and class imbalance distribution of in-the-wild datasets without deteriorating feature learning remains a challenge. To solve these problems, we propose Combating Uncertainty and Class Imbalance Network (CUCN) to simultaneously suppress uncertainty and solve the class imbalance problem of facial expression recognition. First, the feature extraction network of CUCN will extract an uncertain feature and the feature used to recognize the expressions of each sample. Just like the data enhancement strategy Mixup, we will mix these two features in the same batch of samples. Then, we add a weight to each expression class as we input the images. For classes with a large number of samples, the model suppresses their learning, and conversely, for classes with a small number of samples, it suppresses their learning. And we embed Convolutional Block Attention Module (CBAM) in the model, which adaptively modifies our extracted features to better learn contextual features and improve the accuracy of the model.

Overall, the main contributions of our work can be summarized as follows:
\begin{itemize}
\item We have proposed CUCN to solve the uncertainty of data and class imbalance which can maintain the accuracy of the model while equalizing the accuracy of each expression class.
\item We have added the weight mechanism and attention module CBAM to CUCN, which can balance the recognition accuracy of each expression.
\item We have conducted extensive experiments on widely-used FER benchmarks, including AffectNet and RAF-DB to demonstrate the effectiveness of our CUCN framework.
\end{itemize}

\section{Method}

\subsection{Overview}

In order to reduce the interference of the uncertainty of facial expression dataset on the model and the bias of class imbalance on the prediction of each expression, we propose a simple and effective framework, namely, the Combating Uncertainty and Class Imbalance Network (CUCN), and this work is an improvement made on RUL\cite{zhang2021relative}. Its general structure is shown in Fig.~\ref{fig2}. For the input pictures, we first apply a weight to them by class. The class with a large number of samples applies a larger weight, and the class with a small number of samples applies a smaller weight. The model will pay more attention to those samples with a small weight, so as to suppress the problem of class imbalance. We use a residual network as our backbone network, this classic network can help our model better learn features. In order to better learn fine-grained features, we add CBAM to it, which brings negligible additional overhead to the network and will bring a certain performance improvement to the model. The backbone network will learn two features for samples, which we call classification features and uncertainty features. Through a strategy similar to the data enhancement Mixup operation, we mix the classification features and uncertainty features between samples to enhance those samples with stronger uncertainty. Finally, using the mixed features to classify through a full connection layer, we can get good results in terms of uncertainty and class imbalance.

\subsection{Convolutional Block Attention Module}

\begin{figure}[htbp]
\centerline{\includegraphics[scale=0.7]{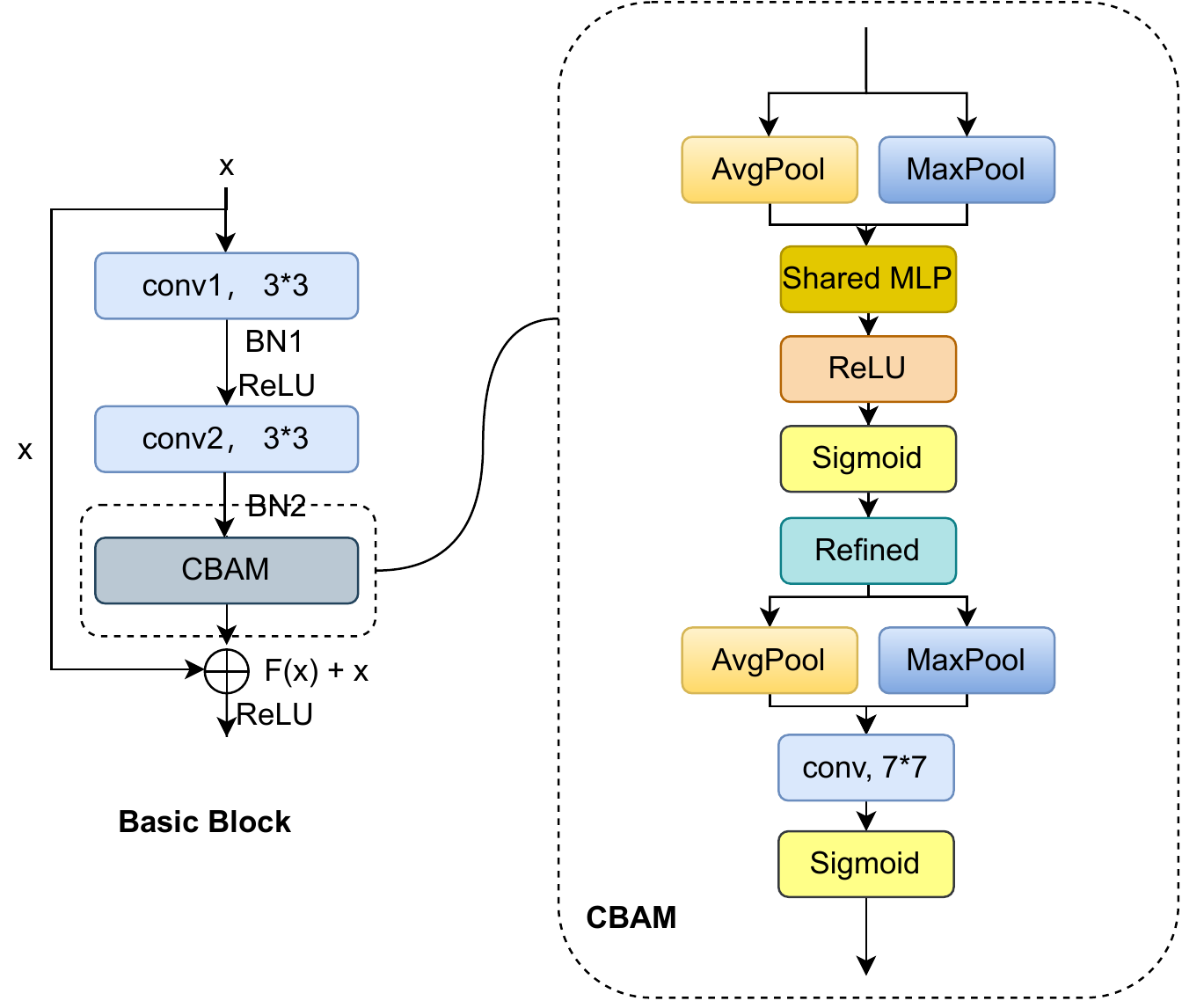}}
\caption{The module on the left is the Basic Block, and the module on the right is the embedded CBAM. The upper half is the CA module and the lower half is the SA module. They are arranged in sequence.}
\label{fig3}
\end{figure}

The inclusion of a weighting mechanism in the model alleviates the problem of class imbalance, but it has also brought the problem of declining accuracy. Therefore, we introduced the Convolutional Block Attention Module (CBAM)\cite{woo2018cbam}. Because of the characteristics of the residual network, it can extract features at multiple scales, so we put this module in the shortcut connection (Basic Block) of the residual network to help information at various scales to flow efficiently in the network.  The module consists of a channel attention module (CA) and a spatial attention module (SA). Since our method requires not only extracting recognition features and uncertainty features from the samples but also mixing features between samples, this module enhances the representation capability of our model to some extent and improves the accuracy. The working mechanism of CBAM in our model is shown in Fig.~\ref{fig3}.

\subsection{Loss Function}

In this paper, we use the most commonly used cross-entropy loss function in dealing with the classification task and make some modifications to the traditional cross-entropy loss (Cross Entropy, CE) function to make it more suitable for our model. We apply the previously designed class weights to the cross-entropy loss function, and the cross-entropy loss function after adding the weights is:
\begin{equation}
L_{weight}=-\frac{1}{N}W_{n} \sum_{n=0}^{N-1} \log_{}{\left ( p_{n} \right )} 
\end{equation}

Among them, \textit{N} is the number of images, and $p_{n}$ is the maximum probability value of the predicted result for the \textit{n-th} sample.$W_{n}$ is the weight coefficient, which is a hyperparameter we set according to the degree of class imbalance in the dataset. According to this weight, the network model can be made to pay more attention to the learning of small samples during the training process. For uncertainty, we follow the cumulative loss in RUL, whose formula is as follows:
\begin{equation}
L_{total}=-\frac{1}{N}
\sum_{i,j}^{N}\left(
\log_{}
{\frac{e^{W_{y_{i}}\tilde{\mu} } }{ {\textstyle \sum_{c}^{C}e^{W_{c}\tilde{\mu} }} }}+
\log_{}
{\frac{e^{W_{y_{j}}\tilde{\mu} } }{ {\textstyle \sum_{c}^{C}e^{W_{c}\tilde{\mu} }} }}
\right ) 
\end{equation}

Where $\tilde{\mu}$ is the mixed feature, the loss function adds up the loss of recognizing $\tilde{\mu}$ as class i and as class j. $y_{i}$, $y_{j}$ means $label_{i}$, $label_{j}$, $W_{c}$ is the \textit{c-th} classifier and \textit{C} means the total number of expression classes. The loss function forces the model to recognize two expressions equally from the mixed features. When mixing two facial features, there will be a relatively easy facial feature for expression recognition, and the other will be relatively hard. As a result, hard samples have to take up a large amount of the mixed feature $\tilde{\mu}$.

\section{Experiment}

\subsection{Implementation Details}

\begin{table}[htbp]
\caption{Comparison with Other State-of-the-Art Result on Different Datasets}
\begin{center}
\resizebox{85mm}{!}{
\begin{tabular}{|c|c|c|c|c|c|} 
\hline
\multicolumn{3}{|c|}{\textbf{RAF-DB}}        & \multicolumn{3}{c|}{\textbf{AffectNet}}       \\ 
\hline
\textbf{Method}      & \textbf{Year} & \textbf{Accuracy} (\%)  & \textbf{Method}      & \textbf{Year} & \textbf{Accuracy} (\%)   \\ 
\hline
SCN\cite{Wang_2020_CVPR}         & 2020 & 87.03          & SCN\cite{Wang_2020_CVPR}         & 2020 & 60.23           \\ 
\hline
DMUE\cite{she2021dive}        & 2021 & \textbf{88.76} & DMUE\cite{she2021dive}        & 2021 & 62.84           \\ 
\hline
DACL\cite{farzaneh2021facial}        & 2021 & 87.78          & T21DST\cite{xie2021triplet}      & 2021 & 60.12           \\ 
\hline
VTFF\cite{ma2021facial}        & 2021 & 88.14          & WSFER\cite{zhang2021weakly}       & 2021 & 60.04           \\ 
\hline
Ad-core\cite{fard2022ad}     & 2022 & 86.96          & Ad-core\cite{fard2022ad}     & 2022 & 63.36           \\ 
\hline
CUCN (ours) & 2022 & 88.66          & CUCN (ours) & 2022 & \textbf{63.77}  \\
\hline
\end{tabular}}
\label{tab1}
\end{center}
\end{table}

\begin{figure}[htbp]
\centerline{\includegraphics[scale=0.5]{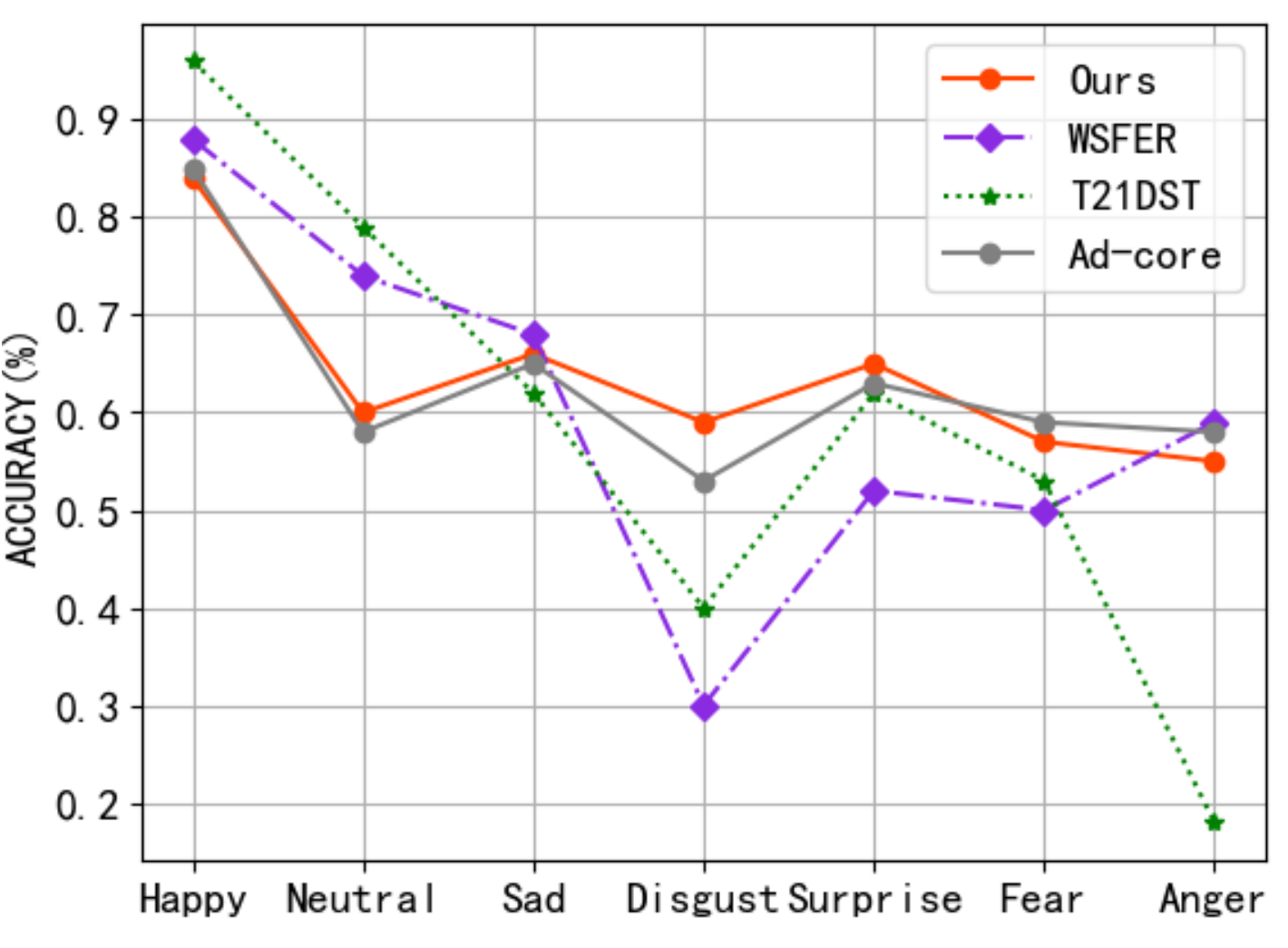}}
\caption{A Line chart comparing the accuracy of each expression with other methods on AffectNet dataset.}
\label{fig4}
\end{figure}

\begin{figure*}[htbp]
\centerline{\includegraphics[scale=0.5]{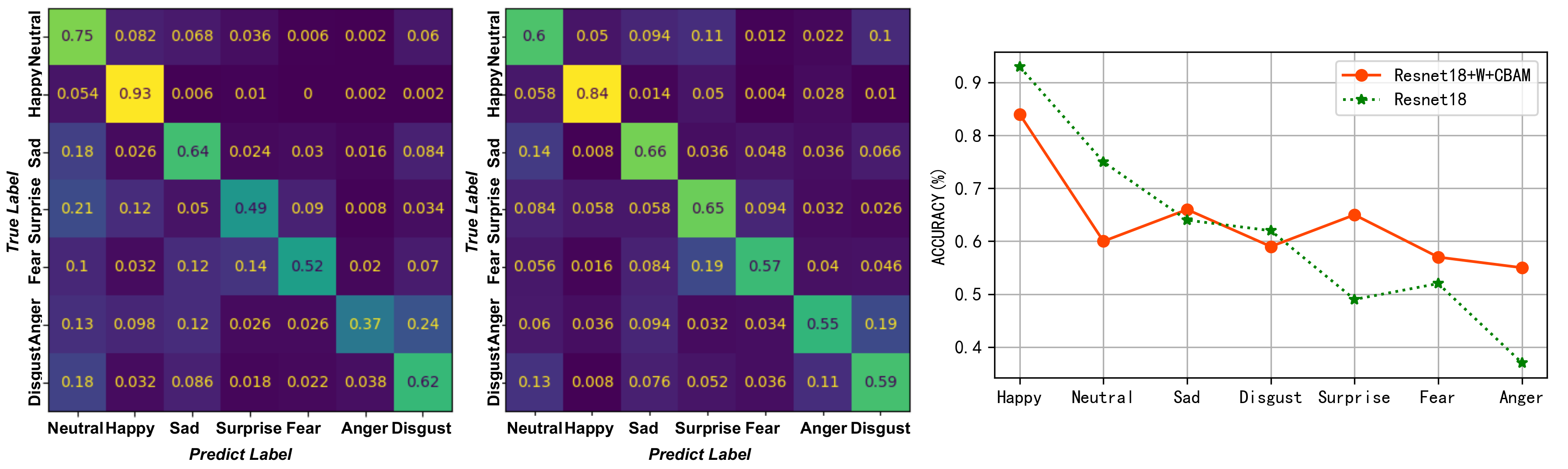}}
\caption{The confusion matrix on the left is before adding weights, the right one is the result after adding weights and CBAM, and the line chart is their accuracy comparison.}
\label{fig5}
\end{figure*}

We used \textbf{RAF-DB}\cite{li2017reliable} and \textbf{AffectNet}\cite{mollahosseini2017affectnet} as our experimental datasets. Given a batch of training set images, we first resize them to \(224\times 224\) pixels, then perform enhancement operations such as random flipping (the test set is only resized, not enhanced), and then manually set the class weights and put them into ResNet18, which is used as a backbone network after pre-training on Ms-Celeb-1M face recognition dataset. Finally, we extract the discriminative features and uncertain features learned by the model from the penultimate layer of the residual network for mixing. The model is trained in an end-to-end manner using a single GTX 2080ti GPU. Our hyperparameter settings are slightly different for different datasets. Using AffectNet as an example, we trained 60 epochs with a batch size of 256, used Adam optimizer with a weight attenuation of 0.0001, initialized the learning rate to 0.0001, and initialized the full connection learning rate for classification to 0.0005. We also use the ExponentialLR learning rate scheduler with a gamma of 0.9 to gradually reduce the learning rate of each epoch.

\subsection{Comparison with State-of-the-art}

In Table~\ref{tab1}, we show the results of comparing our method with recent publicly available methods on the RAF-DB and AffectNet datasets, and we convert their results to accuracy for a fair comparison. The table shows that the accuracy of our method reaches a level comparable to current state-of-the-art methods and is only slightly lower than DMUE [24] on the RAF-DB dataset.

In Fig.~\ref{fig4} we show the results of comparing the accuracy of our method with some of their other methods on the AffectNet dataset in seven classes of expressions (the results are given by their papers; we did not compare for methods that did not give corresponding data or confusion matrices), and since most of their methods included contempt, we excluded the contempt data for fairness, and it can be clearly seen that our method is more balanced than theirs in terms of accuracy across expression class.

\subsection{Ablation Study}

To demonstrate our contribution to class imbalance, we present our confusion matrix and line graphs before and after adding class weights on AffectNet. As shown in Fig.~\ref{fig5}, after adding the weighting mechanism, the classification results appear more balanced. Also, we conducted a series of ablation experiments to verify the effectiveness of our module. We first used only the backbone network ResNet18 to extract features and achieved 88.79\% on RAF-DB, which shows that our backbone network is effective in suppressing uncertainty. Since the AffectNet dataset is larger and class imbalance is more pronounced, we validate the effectiveness of our weighting module in suppressing class imbalance on  AffectNet. By replacing the pre-training model and the backbone network, we compare the situation before and after adding the weight module and the CBAM module. Through the comparison, we find that the highest and lowest recognition rate intervals are significantly smaller after adding weights than before adding weights, which indicates that our method is effective. By comparing adding only weights and adding weights and CBAM, we found that the accuracy of the model also improved to some extent. The results are given in Table~\ref{tab2}, where '*' represents the official pre-trained model provided by Pytorch, and '\#' represents the model pre-trained on Ms-Celeb-1M.

\begin{table}
\caption{Comparison Before and After Adding Weight and CBAM with Different Pre-trained Model and Resnet}
\begin{center}
\resizebox{85mm}{!}{
\begin{tabular}{|c|c|c|c|c|c|c|c|c|c|} 
\hline
\textbf{Method}            & \textbf{Dataset}   & \textbf{Max Accu} (\%) & \textbf{Min Accu} (\%) & \textbf{Accuracy} (\%)  \\ 
\hline
ResNet18\#        & AffectNet & 93         & 37         & 61.57          \\ 
\hline
ResNet18\#+W      & AffectNet & 84        & 55         & 63.34          \\ 
\hline
ResNet18\#+W+CBAM & AffectNet & 84         & 59         & 63.77          \\ 
\hline
ResNet18*         & AffectNet & 93        & 37         & 60.14          \\ 
\hline
ResNet18*+W+CBAM  & AffectNet & 82         & 55         & 62.71          \\ 
\hline
ResNet50*         & AffectNet & 92         & 34         & 60.28          \\ 
\hline
ResNet50*+W+CBAM  & AffectNet & 83         & 53         & 63.29          \\
\hline
\end{tabular}}
\label{tab2}
\end{center}
\end{table}

\section{Conclusion}

In this paper, we have proposed a simple and effective framework named CUCN, whose backbone network adopts a modified ResNet18.We have added a weighting module and embeded a CBAM module with a mixed of uncertain features and recognition features extracted from the backbone network. The experimental results have shown that our method can well suppress the problems of class imbalance and uncertainty in large-scale expression datasets, make the recognition accuracy of each expression class more balanced, and the average accuracy of the model is equivalent to the mainstream methods.

\section*{Acknowledgment}

This work was supported in part by the National Natural Science Foundation of China under Grant 61372137, in part by the Natural Science Foundation of Anhui Province under Grant 1908085MF209.

%\section*{References}

%\begin{thebibliography}{99}
\bibliographystyle{IEEEtran}
\bibliography{IEEEabrv,mybib}
%\end{thebibliography}

\end{document}